\documentclass{article} 
\usepackage{aloe2022,times}

\author{Luca Grillotti \& Antoine Cully \\
Imperial College London\\
London, UK \\
\texttt{\{luca.grillotti16, a.cully\}@imperial.ac.uk} \\
}

\usepackage{amsmath,amsfonts,bm}

\def\eqref#1{equation~\ref{#1}}

\def\1{\bm{1}}

\DeclareMathAlphabet{\mathsfit}{\encodingdefault}{\sfdefault}{m}{sl}
\SetMathAlphabet{\mathsfit}{bold}{\encodingdefault}{\sfdefault}{bx}{n}

\usepackage{graphicx}
\usepackage{hyperref}
\usepackage{url}
\usepackage{amsmath}
\usepackage{subcaption}
\usepackage{dirtytalk}

\usepackage{tikz}
\usepackage{import}
\usepackage{xifthen}
\usepackage{pdfpages}
\usepackage{transparent}
\usepackage{layouts}

\usetikzlibrary{bayesnet}
\usetikzlibrary{arrows}

\usepackage{booktabs} 
\usepackage{threeparttable}

\usepackage{algorithm}[H] 
\usepackage[noend]{algpseudocode}
\newbox\statebox
\newcommand{\myState}[1]{%
    \setbox\statebox=\vbox{#1}%
    \edef\thealgruleheight{\dimexpr \the\ht\statebox+1pt\relax}%
    \edef\thealgruledepth{\dimexpr \the\dp\statebox+1pt\relax}%
    \ifdim\thealgruleheight<.75\baselineskip
        \def\thealgruleheight{\dimexpr .75\baselineskip+1pt\relax}%
    \fi
    \ifdim\thealgruledepth<.25\baselineskip
        \def\thealgruledepth{\dimexpr .25\baselineskip+1pt\relax}%
    \fi
    \State #1%
    \def\thealgruleheight{\dimexpr .75\baselineskip+1pt\relax}%
    \def\thealgruledepth{\dimexpr .25\baselineskip+1pt\relax}%
}

\title{Discovering Unsupervised Behaviours from Full-State Trajectories}

\newcommand{\name}{AURORA}

\renewcommand{\vec}[1]{{\boldsymbol{{#1}}}} 

\newcommand{\B}{\mathcal{B}}

\newcommand{\cC}{\mathcal{C}}
\newcommand{\cT}{\mathcal{T}}
\newcommand{\cX}{\mathcal{X}}
\newcommand{\cB}{\mathcal{B}}
\newcommand{\cO}{\mathcal{O}}

\newcommand{\card}[1]{\lvert #1 \rvert}
\newcommand{\abs}[1]{\lvert #1 \rvert}

\newcommand{\targetscC}{N_{\cC}^{target}}

\newcommand{\encoder}{\mathcal{E}}
\newcommand{\paramEncoder}{\vec\eta_{\mathcal{E}}}

\newcommand{\bdtaskspace}{\B_\cT}
\newcommand{\bdtask}{\vec b_{\cT}}
\newcommand{\bdprop}{\vec b_{\encoder}}

\algnewcommand\algorithmicforeach{\textbf{for each}}
\algdef{S}[FOR]{ForEach}[1]{\algorithmicforeach\ #1\ \algorithmicdo}

\newcommand{\states}{\vec s_{1:T}}
\newcommand{\obs}{\states}

\newcommand{\entropy}[1]{{H\left({{#1}}\right)}} 
\newcommand{\entropyCond}[2]{H\left({#1} \mid {#2}\right)} 

\newcommand{\aurora}{AURORA}

\usepackage{pifont}

\iclrfinalcopy 

\begin{document}

\maketitle

\begin{abstract}
Improving open-ended learning capabilities is a promising approach to enable robots to face the unbounded complexity of the real-world. 
%
%
%
Among existing methods, the ability of Quality-Diversity algorithms to generate large collections of diverse and high-performing skills is instrumental in this context. 
However, most of those algorithms rely on a hand-coded behavioural descriptor to characterise the diversity, hence requiring prior knowledge about the considered tasks. 
In this work, we propose an additional analysis of Autonomous Robots Realising their Abilities; a Quality-Diversity algorithm that autonomously finds behavioural characterisations.
We evaluate this approach on a simulated robotic environment, where the robot has to autonomously discover its abilities from its full-state trajectories.
All algorithms were applied to three tasks: navigation, moving forward with a high velocity, and performing half-rolls. 
The experimental results show that the algorithm under study discovers autonomously collections of solutions that are diverse with respect to all tasks.
More specifically, the analysed approach autonomously finds policies that make the robot move to diverse positions, but also utilise its legs in diverse ways, and even perform half-rolls.
\end{abstract}

\section{Introduction}






One of the motivations of using learning algorithms in robotics is to enable robots to discover on their own their abilities. 
Ideally, a robot should be able to discover on its own how to manipulate new objects~\citep{johns2016deep} or how to adapt its behaviours when facing an unseen situation like a mechanical damage~\citep{Cully2014robotsanimals}. 
However, despite many impressive breakthroughs in learning algorithms for robotics applications during the last decade~\citep{akkaya2019solving, hwangbo2019learning}, these methods still require a significant amount of engineering to become effective, for instance, to define reward functions, state definitions, or characterise the expected behaviours. 

Finding all possible behaviours for a robot is an open-ended process, as the set of all achievable robot behaviours is unboundedly complex.
There are plenty of different ways to execute each type of behaviour.
For instance, for moving forward, rotating or jumping, the robot can move at different speeds, and use different locomotion gaits. 
Supplementary behaviours can emerge from interactions with objects and with other agents.

An attempt to discover autonomously the behaviours of a robot has been proposed with the AURORA algorithm~\citep{Cully2019,grillotti2021unsupervised}. 
This algorithm leverages the creativity of Quality-Diversity (QD) optimisation algorithms to generate a collection of diverse and high-performing behaviours, called a behavioural repertoire. 
QD algorithms usually require the definition of a manually-defined Behavioural Descriptor (BD) to characterise the different types of behaviours contained in the repertoire. 
Instead, AURORA uses dimensionality-reduction techniques to perform the behavioural characterisation in an unsupervised manner. 
The resulting algorithms enable robots to autonomously discover a large diversity of behaviours without requiring a user to define a performance function or a behavioural descriptor. 
Nevertheless, in prior work, AURORA has only been used to learn behavioural characterisations from \textit{hand-defined} sensory data.
For example, \citet{grillotti2021unsupervised} use as sensory data a robot picture taken at the end of an episode from an appropriate camera angle.
In that case, a consequent amount of prior information is provided to the algorithm: only the end of the episode should be considered (instead of the full trajectory), and only the given camera angles should be used (instead of other camera angles or joints positions).



%
In this work, we propose an extended analysis of \aurora{}, where behavioural characterisations are learned based on the complete trajectory of the robot raw states.
In that sense, no prior information is injected in the sensory data.
We evaluated \aurora{} on a simulated hexapod robot with a neural network controller on three tasks: navigation, moving forward with a high velocity, and performing half-rolls.
Furthermore, we compare \aurora{} to several baselines, all of which use hand-coded BDs. 
In most cases, the collections obtained via \aurora{} present more diversity than those obtained from hand-coded QD algorithms.
In particular, \aurora{} automatically discovers behaviours that make the robot navigate to diverse positions, while using its legs in diverse ways, and perform half-rolls.

\section{Background}

\subsection{Quality-Diversity Algorithms}

\label{sec:background_qd}

Quality-Diversity (QD) algorithms are a subclass of evolutionary algorithms that aim at finding a container of both diverse and high-performing policies.
In addition to standard evolutionary algorithms, QD algorithms consider a Behavioural Descriptor (BD), which is a low-dimensional vector that characterises the behaviour of a policy.
QD algorithms use the BDs to quantify the novelty of a policy with respect to the solutions already in the container. 
The container of policies is filled in an iterative manner, following these steps: (1) policies are selected from the container and undergo random variations (e.g. mutations, cross-overs); (2) their performance score and BD are evaluated; and (3) we try to add them back to the container.
If they are novel enough compared to solutions that are already in the container, they are added to the container.
If they are better than similar policies in the container, they replace these policies.

\subsection{Autonomous Robots Realising their Abilities (AURORA)}

\label{subsec:unsupervised_behaviours}

Quality diversity algorithms are a promising tool to generate a large diversity of behaviours in robotics. However, the definition of the BD space might not always be straightforward when the robot or its capabilities are unknown. AURORA is a QD algorithm designed to discover the abilities of robots, by maximising the diversity of behaviours contained in the repertoire in an unsupervised manner. AURORA automatically defines the BD by encoding the high-dimensional sensory data collected by the robot during an episode.
The encoding can be achieved using any dimensionality reduction technique, such as, Auto-Encoder (AE) \citep{grillotti2021unsupervised} or Principal Component Analysis \citep{Cully2019}.

To generate a behavioural repertoire, AURORA alternates between QD phases and Encoder phases.
During the QD phase, policies undergo the same process as in standard QD algorithms: selection from the container, evaluation, and attempt to add to the container. However, the evaluation is performed in a slightly different way: instead of a low-dimensional BD the QD task returns high-dimensional sensory data, that is then encoded using the dimensionality reduction technique. The latent encoding of the sensory data is used as the BD for the rest of the QD phrase.

During the Encoder phase, the dimensionality reduction structure (e.g. the AE) is trained using the high-dimensional data from all the policies present in the container.
Once the encoder is trained, the unsupervised BDs are recomputed with the new encoder for all policies present in the container.
    
Alternating these two phases enables AURORA to progressively build its behavioural repertoires by discovering new solutions, while refining its encoding of the high-dimensional sensory data generated by the robot every time new solutions are added to the archive.
In this work, the sensory data collected by the robot corresponds to the full-state trajectory of the robot.


\section{Learning Diverse Behaviours from Full-State Trajectories}

In this section, we provide a theoretical analysis of the AURORA algorithm, to support its ability to generate a large collection of diverse behaviours in an unsupervised manner.
%
%
In essence, we aim to find a container of policies exhibiting diverse behaviours.
We consider that a \textit{behaviour} is defined as the full-state trajectory: $\states$.
Then the overall problem of finding diverse behaviours $\left(\states^{(i)}\right)_{i}$ can be expressed as an entropy maximisation problem. 
Considering the behaviour $\states$ as a random variable, we aim to maximise the entropy of $\states$, written $\entropy{\states}$.
In the following sections we will explain (1) why it can be ineffective to maximise directly the diversity of full-state trajectories $\states$; and (2) how AURORA can be used to maximise indirectly this diversity.

\subsection{Problem with High-dimensional Behavioural Descriptors}

Maximising the entropy of full-state trajectories $\entropy{\states}$ is theoretically possible by using a QD algorithm with $\states$ as a Behavioural Descriptor (BD).
However, it is computationally inefficient to use QD algorithms with high-dimensional BDs. 

Indeed, as the dimensionality of that problem is high, a consequent number of samples of $\states$ is required to optimise $\entropy{\states}$.
So a significant number of policies is needed in the container.
However, QD algorithms become less efficient when the number of policies increases.
When the number of policies in the container increases, the selection pressure of the container decreases, which means that the policies from the container are mostly less likely to be selected \citep{Cully2018QDFramework}.

Also, in QD algorithms, unstructured containers always rely on the k-nearest neighbours algorithm to compute the novelty of a policy.
And the complexity of that algorithm can be $O\left( \dim{\cB}\log \abs{\cC}  \right)$ in average \citep{bentley1975multidimensional}, and $O\left( \dim{\cB} \abs{\cC}  \right)$ in the worst case  ($\cB$ refers to the BD space, and $\abs{\cC}$ is the number of policies in the container $\cC$).
Hence, the computational complexity of one QD generation increases linearly with dimensionality of the BD $\dim{\cB}$.
That is why the overall procedure of adding policies to the container at each QD generation may become too computationally expensive if the dimension of $\cB$ increases significantly.



\subsection{Learning Low-dimensional Descriptors from High-dimensional State Trajectories}

\label{sec:learning-low-from-high}

\newcommand{\mi}[2]{I\left( #1 , #2 \right)}

\begin{figure}
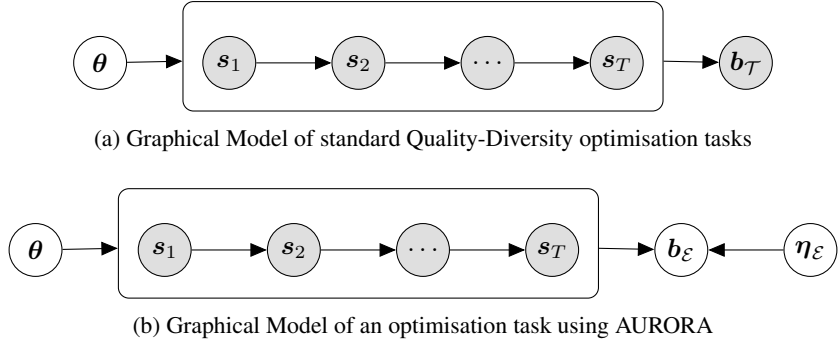

  \centering
      \begin{subfigure}[b]{\textwidth}
         \centering
              \tikz{
 \node[latent] (gen) {$\vec\theta$};%
 \node[obs,right=of gen] (s1) {$\vec s_1$}; %
 \node[obs,right=of s1] (s2) {$\vec s_2$}; %
 \node[obs,right=of s2] (sdots) {$\ldots$}; %
 \node[obs,right=of sdots] (sT) {$\vec s_T$}; %

 \node[obs, right=of sT] (bt) {$\bdtask$};
 \plate [inner sep=.25cm,yshift=.2cm] {plate1} {(s1)(s2)(sdots)(sT)} {}; %

 \edge {s1} {s2}
 \edge {s2} {sdots}
 \edge {sdots} {sT}
 \edge {gen} {plate1}
 \edge {plate1} {bt}
 }
         \caption{Graphical Model of standard Quality-Diversity optimisation tasks}
         \label{subfig:QD_diagram}
         \end{subfigure}
         
\par\bigskip

\begin{subfigure}[b]{\textwidth}
    \centering
              \tikz{
 \node[latent] (gen) {$\vec\theta$};%
 \node[obs,right=of gen] (s1) {$\vec s_1$}; %
 \node[obs,right=of s1] (s2) {$\vec s_2$}; %
 \node[obs,right=of s2] (sdots) {$\ldots$}; %
 \node[obs,right=of sdots] (sT) {$\vec s_T$}; %
 \node[latent, right=of sT] (bt) {$ \bdprop$};
 \node[latent, right=of bt] (encoder) {$\paramEncoder{}$};
 \plate [inner sep=.25cm,yshift=.2cm] {plate1} {(s1)(s2)(sdots)(sT)} {}; %

 \edge {s1} {s2}
 \edge {s2} {sdots}
 \edge {sdots} {sT}
 \edge {gen} {plate1}
 \edge {plate1} {bt}
 \edge {encoder} {bt}
 }
    \caption{Graphical Model of an optimisation task using AURORA}
    \label{fig:diagramAurora}
\end{subfigure}

\par\bigskip

 \caption[Graphical Models of several Quality-Diversity algorithms under study]{
 Graphical Models of the Quality-Diversity optimisation algorithms in the standard case (\ref{subfig:QD_diagram}) and using AURORA (\ref{fig:diagramAurora}). 
 $\vec \theta$ corresponds to the parameters of the agent's policy. 
 At each time-step $t$, $\vec s_t$ represent the state of the agent. 
 $\bdtask$ represents the task behavioural descriptor based on the states of the robot; $ \bdprop$ is a behavioural descriptor based on the successive states of the robot $\states$.
 Finally, $\paramEncoder{}$ corresponds to the parameters of the encoder $\mathcal{E}$ used to reduce the dimensionality of the state trajectories $\states$.
 }
 \label{fig:graphical_model}
\end{figure}

The encoder update phase of AURORA optimises the parameters $\paramEncoder$ of the encoder $\encoder$; this encoder is then used to define a low-dimensional BD $\bdprop$ from state trajectories: $\bdprop = \encoder(\obs)$. 
Figure~\ref{fig:graphical_model} summarises the dependencies between the different variables considered in standard QD algorithms and AURORA.
The mutual information $\mi{\cdot}{\cdot}$ between $\obs$ and $\bdprop$ can be expressed in two symmetrical ways:
\begin{align}
\mi{\obs}{\bdprop} &= \entropy{\obs} - \entropyCond{\obs}{\bdprop} \label{eq:mi_obs} \\ 
\mi{\obs}{\bdprop} &= \entropy{\bdprop} - \entropyCond{\bdprop}{\obs} \label{eq:mi_bdprop}
\end{align}

As $\bdprop$ is a function of $\obs$, we have: $\entropyCond{\bdprop}{\obs} = 0$. 
Then, from equations~\ref{eq:mi_obs} and~\ref{eq:mi_bdprop} we get:
\begin{align}
    \entropy{\obs} &= \entropy{\bdprop} +  \entropyCond{\obs}{\bdprop} \\
    &= \entropy{\bdprop} + \entropyCond{\obs}{\encoder \left( \obs \right)} \label{eq:entropy_obs_equality}
\end{align}

According to equation~\ref{eq:entropy_obs_equality}, $\entropy{\obs}$ can be decomposed into two components:
\begin{itemize}
    \item The entropy on the BD space $\entropy{\bdprop}$, which can be maximised via a QD algorithm.
    \item The conditional entropy of the state trajectories with respect to their encoding $\entropyCond{\obs}{\encoder \left( \obs \right)}$. 
    This term quantifies the information that is lost between $\obs$ and its encoding $\bdprop = \encoder \left( \obs \right)$.
\end{itemize}

Inequality~\ref{eq:entropy_obs_equality} gives a lower bound on the entropy of state trajectories:
\begin{align}
    \entropy{\obs} \geq \entropy{\bdprop} \label{eq:lower-bound-obs}
\end{align}
And Fano's inequality \citep[p.~38]{cover2012elements} gives a bounding on the differences between those entropies:
\begin{align}
    \entropy{\obs} - \entropy{\bdprop} &= \entropyCond{\obs}{\encoder \left( \obs \right)} \\ 
    \entropy{\obs} - \entropy{\bdprop} &\leq \entropy{e} + P(e) \log \card{\cX_{\cO^T}} \label{eq:inequality_differences}
\end{align}
where $P(e) = P(\obs \neq \widehat{\obs})$, $\widehat{\obs}$ represents a reconstruction of the state trajectories $\obs$ from the encoding $\encoder(\obs)=\bdprop$, and $\cX_{\cO^T}$ represents the discretised support of the state trajectories $\obs$.

As explained in section~\ref{subsec:unsupervised_behaviours}, AURORA alternates between two phases: an \textit{encoder update} phase, and a \textit{QD} phase.
%
%
During the encoder update phase, we minimise the difference $\entropy{\obs} - \entropy{\bdprop}=\entropy{\obs} - \entropy{\encoder(\obs)}$ by training the encoder $\encoder$.
Indeed, training $\encoder$ will reduce the probability of reconstruction error $P(e)$, with the intention of lessening the difference $\entropy{\obs} - \entropy{\bdprop}$ (see inequality~\ref{eq:inequality_differences}).
During the QD phase, now that the difference between both sides of inequality~\ref{eq:lower-bound-obs} is minimised, we increase the lower bound of $\entropy{\obs}$, namely $\entropy{\bdprop}$.
As $\bdprop$ is a low-dimensional BD, maximising $\entropy{\bdprop}$ can be intuitively achieved by applying a standard Quality-Diversity algorithm, with $\bdprop$ as BD.

In the end, the two phases of AURORA successively (1)~train the encoder to minimise the difference between the entropy of successive states $\entropy{\states}$ and its lower bound $\entropy{\encoder (\states)}$, and~(2) use QD iterations, to maximise the lower bound $\entropy{\bdprop} = \entropy{\encoder (\states)}$.
This intuitively explains why AURORA can be used to maximise the diversity of full-state trajectories $\states$.

\section{Related Works}

The problem of autonomously finding diverse behaviours has been addressed with Mutual-Information (MI) maximisation approaches \citep{gregor2016variational, eysenbach2018diversity, sharma2019dynamics}.
Such methods aim to maximise the mutual information between the states explored by the robot, and a latent variable representing the behaviour.
Those MI-based approaches then return a single policy, which is conditioned on the behaviour latent variable.
That differs from QD algorithms, which return a container of different policies.
Also, MI-maximisation algorithms do not necessarily require the definition of a BD. For example, in the work of \citet{eysenbach2018diversity}, diverse robotic behaviours are learned from full robot states.
Similarly to AURORA, when the robot states are described with high-dimensional images, an encoder can be used to compute tractable low-dimensional representations \citep{liu2021apt, liu2021aps}.
While sharing the same purpose, QD and MI-maximisation algorithms are fundamentally different in their core design.
Those differences make it challenging to compare the two approaches in a fair manner.

In QD algorithms, selecting the most appropriate definitions for the BD and the performance requires a certain level of expertise. 
For instance, the MAP-Elites algorithm \citep{Mouret2015, Cully2014robotsanimals} is often used with the same hexapod robot, but with two different sets of definitions: 1) the leg duty-factor for the BD, defined as the time proportion each leg is in contact with the ground, with the average speed for the performance~\citep{Cully2014robotsanimals,Cully2018QDFramework} or 2) the final location of the robot after walking during three seconds for the BD, with an orientation error for the performance~\citep{Cully2013,duarte2016evorbc,chatzilygeroudis2018reset}. 
The resulting behavioural repertoires will find different types of applications. 
The first set of definitions is designed for fast damage recovery as it provides a diversity of ways to achieve high-speed gaits, while the second one is designed for navigation tasks as the repertoire enables the robot to move in various directions. 

To avoid the expertise requirements in the behaviour descriptor definition, several methods have been proposed to enable the automatic definition of the behavioural descriptors. 
As described previously, AURORA aims to tackle this problem by using a dimensionality reduction algorithm (a Deep Auto-Encoder~\citep{masci2011stacked}) to project the features of the solutions into a latent space and use this latent space as a behavioural descriptor space. TAXONS~\citep{Paolo2019taxons} is another QD algorithm adopting the same principle; it demonstrated that this approach can scale to camera images to produce large behavioural repertoires for different types of robots. The same concept was also used in the context of Novelty Search prior to AURORA and TAXONS to generate assets for video games with the DeLeNoX algorithm~\citep{Liapis2013}. 
All these methods aim to maximise the diversity of the produced solutions, without specifically considering a downstream task. 

A different direction to automatically define a behavioural descriptor is to use Meta-Learning. The idea is to find the BD definition that maximises the performance of the resulting collection of solutions when used in a downstream task. For instance, \citet{bossens2020learning} considers the damage recovery capabilities provided by a behavioural repertoire as a Meta-Learning objective and search for the linear combination of a pre-defined set of behaviour descriptors that will maximise this objective. A related approach proposed by \citet{meyerson2016learning} uses a set of training tasks to learn what would be the most appropriate BD definition to solve a "test task".

\section{Experimental Setup}

\subsection{Agent: Neural-network controlled hexapod}

In all our experiments, our agent consists of a simulated hexapod robot controlled via a neural network controller.
The hexapod robot presents 18 controllable joints (3 per leg). 
Each leg $l$ has two Degrees of Freedom: its hip angle $\alpha_l$ and its knee angle $\beta_l$.
The ankle angle is always set to the opposite of the knee angle.

The neural network controller is a Multi-Layer Perceptron with a single hidden layer of size 8.
It takes as input the current joint angles and velocities $(\alpha_l, \beta_l, \dot{\alpha}_l, \dot{\beta}_l)_{1\leq l \leq 6}$ (of dimension 24), and outputs the target values to reach for the joint angles $(\alpha_l, \beta_l)_{1\leq l \leq 6}$ (of dimension 12).
The controller has in total $(24+1)*8 + (8+1)*12 = 308$ independent parameters, operates at a frequency of $50$Hz, while each episode lasts for 3 seconds.
The environment and the controller are deterministic.

\subsection{Dimensionality Reduction Algorithm of \aurora{}}

For each evaluated policy, the state trajectory corresponds to the joint positions, and torso positions and orientations collected at a frequency of 10Hz.
In the end, this represents 18 streams ($(\alpha_i, \beta_i)_{i=1\ldots 6}$, and the positional and rotational coordinates of the torso), with each of those streams containing 30 measurements.
The dimensionality reduction algorithm used in \aurora{} is a reconstruction-based auto-encoder.
The input data is using two 1D convolutional layers, with 128 filters and a kernel of size 3.
Those convolutional layers are followed by one fully-connected linear layer of size 256.
The decoder is made of a fully-connected linear layer of size 256, followed by three 1D deconvolutions with respectively 128, 128 and 18 filters.

The loss function used to train the Encoder corresponds to the mean squared error between the observation passed as input and the reconstruction obtained as output.
The training is performed using the Adam gradient-descent optimiser \citep{kingma2014adam} with $\beta_1=0.9$ and $\beta_2 = 0.999$.
As the encoder needs to be updated less and less frequently as the number of iterations increase, the number of iterations between two encoder updates is linearly increased over time: if the first encoder update happens at iteration 10, then the following encoder updates occur at iterations 30, 60, 100...


\subsection{QD Tasks}

For the hexapod environment, we consider different types of QD Tasks.
A QD task evaluates the diversity and the performance of policies present in the container returned by a QD algorithm.
Each QD task is characterised by a performance score function $\vec \theta \mapsto f(\vec \theta)$, and a BD function $\vec \theta\mapsto \bdtask(\vec \theta)$ (where $\vec \theta$ represents the policy parameters).

We consider three QD tasks with the hexapod agent: Navigation (Nav), Moving Forward (Forw), and Half-roll (Roll).

\subsubsection{Navigation (Nav)}

The Navigation task is inspired from \citet{chatzilygeroudis2018reset} and  \citet{kaushik2020adaptive}.
This task aims at finding a container of controllers reaching diverse final $(x_T, y_T)$ positions in $T=3$ seconds following circular trajectories. 
Hence the hand-coded BD in this case is $\bdtask = (x_T, y_T)$, while the performance $f$ is the final orientation error as defined in the literature~\citep{Cully2013,chatzilygeroudis2018reset}.
    
%

%
%


\subsubsection{Moving Forward (Forw)}

The Moving Forward task is inspired from the work of \citet{Cully2014robotsanimals}, which aims to obtain a container with a variety of ways to move forward at a high velocity.
The hand-coded BD associated to that task $\bdtask$ is the Duty Factor, which
evaluates the proportion of time the each leg is in contact with the ground. 
%
%
The performance score promotes solutions achieving the highest $x$ position at the end of the episode: $f = x_T$.


\subsubsection{Half-roll (Roll)}


Half-roll QD task aims at finding a diversity of ways to perform half-rolls, i.e. behaviours such that the hexapod ends with its back on the floor (where the pitch angle of the torso is approximately equal to $-\frac{\pi}{2}$).
In particular, the behaviours are characterised via the final yaw and roll angle of the torso.
And the performance is measured as the distance between the final pitch angle $\alpha^{pitch}_T$ of the hexapod and $-\frac{\pi}{2}$.
\begin{equation}
    \bdtask = \left(\alpha^{yaw}_{T}, \alpha^{roll}_{T} \right) \quad
    f = - \left\lvert \alpha^{pitch}_{T} - \left(-\frac{\pi}{2}\right) \right\rvert 
\end{equation}

    \subsection{Metric: Coverage per Minimum Performance} 
    
    \label{sec:cov_per_min_fit}
    
    The containers returned by any QD algorithm always result in a trade-off between the diversity and the total performance of the solutions from the container.
    To evaluate the coverage depending on the quality of the solutions, we study the evolution of the coverage given several minimum performance scores.
    With a performance $f_{min}$ and a BD space discretised into a grid, we define the \textit{\say{coverage given minimum performance $f_{min}$}} as: the number of cells filled with policies whose performance is higher than $f_{min}$.
    
    The coverage is calculated by discretising the task Behavioural Descriptor (BD) space~$\bdtaskspace$ into a $50\times 50$ grid.
    The coverage then corresponds the number of grid cells containing at least one policy from the unstructured container.

\subsection{Compared algorithms and variants}

We compare AURORA (see Section~\ref{subsec:unsupervised_behaviours}), to two different variants having hand-defined BDs: Hand-Coded-$x$ (HC-$x$) and Mean Streams (MeS).


%
%

\textbf{Hand-Coded-$x$ (HC-$x$)}: considers a low-dimensional BD that is defined by hand as the BD of the QD task $x$, where $x$ can correspond to any QD task (Navigation, Moving Forward or Half-roll).
The HC variant corresponds to a standard QD algorithm with an unstructured archive~\citep{Cully2018QDFramework} and the mechanism of container size control introduced in previous work \citep{grillotti2021unsupervised}.


\textbf{Mean Streams (MeS)}: uses a BD that encompasses more information from the full-state trajectory than the Hand-Coded variant detailed above.
The BD used by this variant considers the 18 data streams collected by the hexapod, and average each one of those streams to obtain a BD of dimension 18.
%
%
The comparison between this variant and \aurora{} will be appropriate to evaluate the utility of the automatic dimensionality reduction algorithm.

\subsection{Implementation Details}

All our experiments were run for 15{,}000 iterations with a uniform QD selector.
We only use polynomial mutations \citep{deb1999nichedpolynomialmutation} as variation operators with $\eta_m = 10$, and a mutation rate of $0.3$.
The target container size set for all algorithms is set to $\targetscC=5{,}000$ for the Moving Forward and the Half-roll tasks.
In the case of the Navigation Task, we set this number to $\targetscC=1{,}500$ to obtain appropriate comparisons between the different approaches.
To keep the container size around $\targetscC$, we perform a container update every $T_{\cC} = 10$ iterations (as explained in Section~\ref{sec:background_qd}).
%

All our implementation is based on Sferes$_{\text{v2}}$ \citep{Mouret2010} 
and uses on the DART simulator \citep{lee2018dart}, while the auto-encoders are coded and trained using the C++ API of PyTorch \citep{paszke2019pytorch}.
For each QD task, all variants were run for 10 replications.
%

\section{Results}

\begin{figure}
    \centering
    \includegraphics[width=\textwidth]{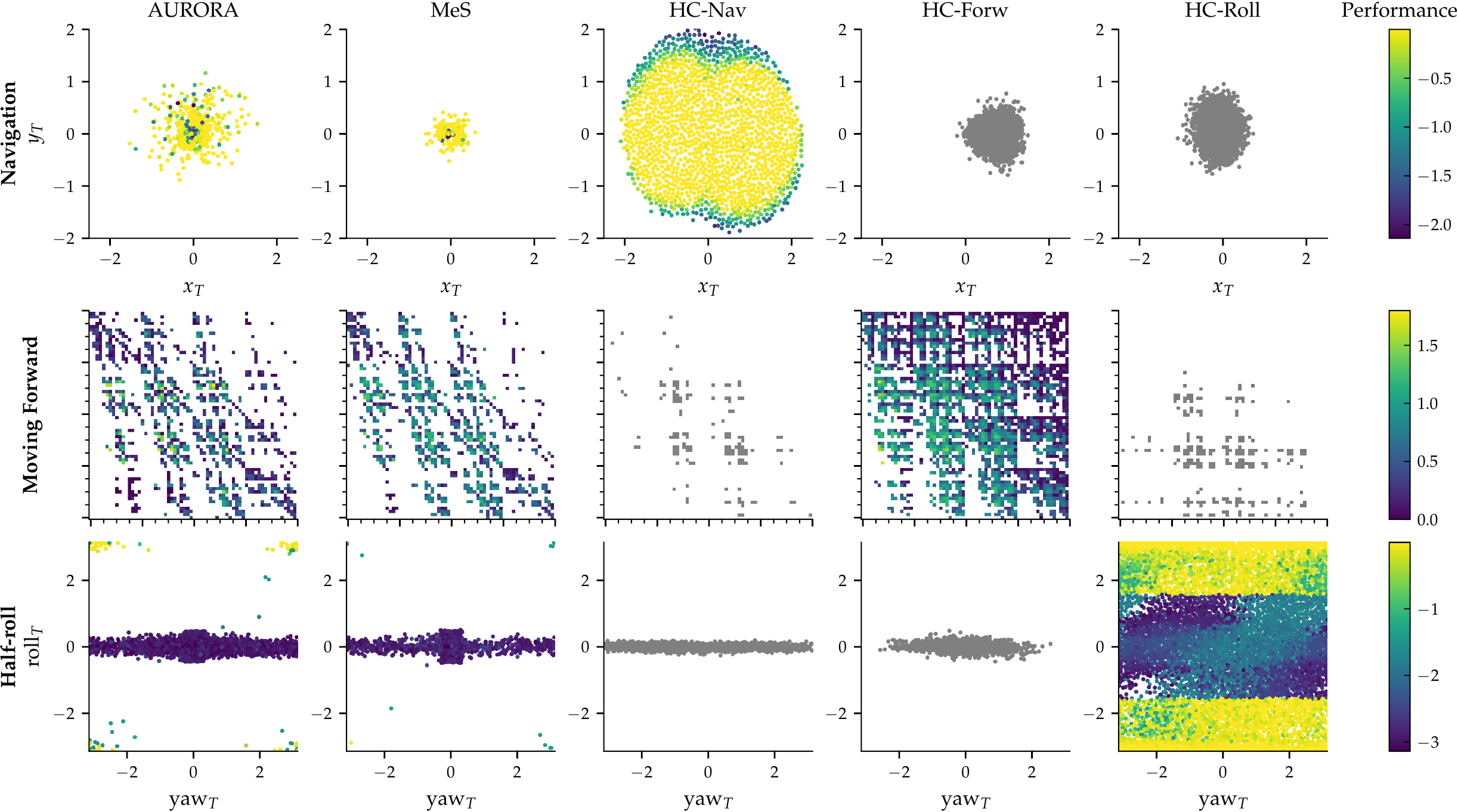}
    \caption{Containers returned by the different variants (columns) for each Quality-Diversity task (rows). 
    Each dot corresponds to one policy present in the container, after projection in the Behavioural Descriptor space $\bdtaskspace$ of the QD task.
    The colour of each dot is representative of the performance score of the policy.
    If the container result from a Hand-Coded variant that is not associated to the task (e.g. the container generated by HC-Nav for the Half-roll task), then the solutions are shown in grey.
    Also, note that the BD space of Duty Factors (second row) has six dimensions, the containers are presented using the same type of grids as in the work of \citet{Cully2014robotsanimals}.
    }
    \label{fig:iclr_01_archives}
\end{figure}

\begin{figure}
    \centering
    \includegraphics[width=\textwidth]{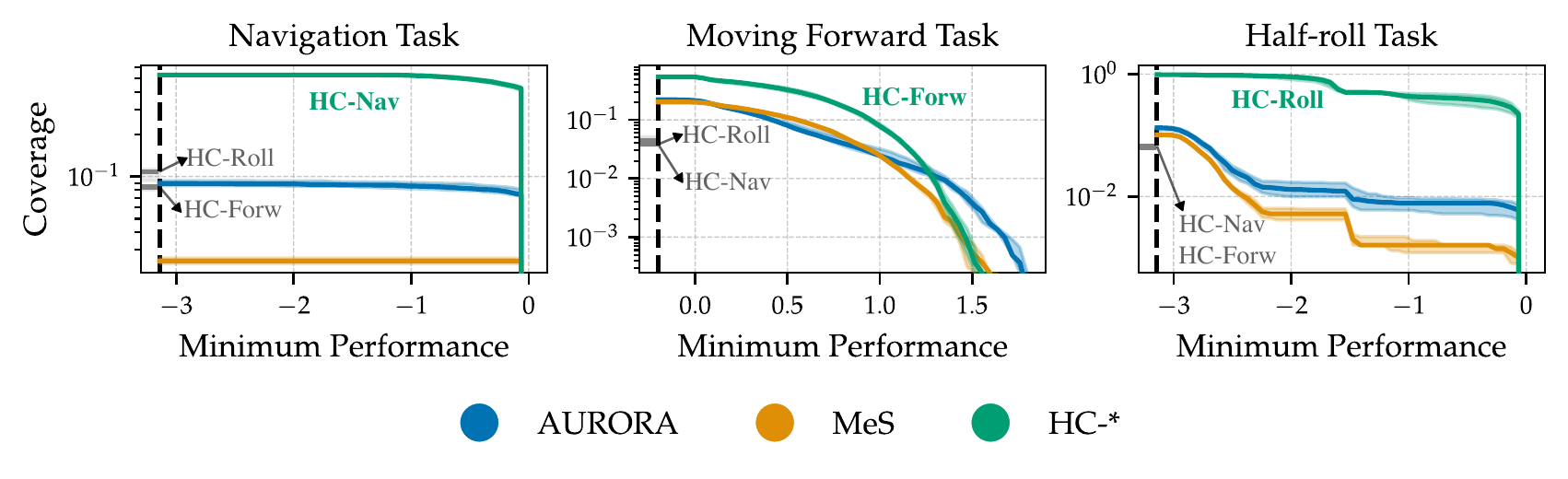}
    \caption{Coverage per minimum performance (presented in section~\ref{sec:cov_per_min_fit}), considered with all downstream tasks under study.
    If the container results from a Hand-Coded variant that is not associated to the task (e.g. the container generated by HC-Forw for the Navigation task), then the performance of solutions are not considered, and only the total coverage is presented.
    The vertical dashed line shows where to read the total coverage (without any minimal performance) achieved by each variant.
    For each variant, the bold line represents the median coverage; and the shaded area indicates the interquartile range. 
    }
    \label{fig:iclr_02_cov_per_min_fit}
\end{figure}

Our objective is to show that AURORA discovers diversity in all directions of the BD space.
To that end, we take the containers returned by the different algorithms, and we project them in the BD spaces associated to each Quality-Diversity (QD) task.
Then we evaluate the achieved coverage in the BD space by considering the coverage for several minimum performance scores.

From Figure~\ref{fig:iclr_02_cov_per_min_fit}, it can be noticed that AURORA systematically obtains a lower total coverage than the variant using the hand-coded BD of the task (e.g. HC-Forw for the Forward task).
Such variants use a BD that is well-adapted to the QD problem to solve; but this kind of BD requires human expertise to be defined.

%
\aurora{} can also be compared to Hand-coded containers after a projection in the BD spaces of other tasks, such as the container from HC-Nav projected in the BD space of the Moving Forward task.
We notice that those kinds of containers (shown in grey in Figs~\ref{fig:iclr_01_archives} and~\ref{fig:iclr_02_cov_per_min_fit}) achieve a coverage that is at most equivalent, if not worse, compared to the containers of AURORA.
For example, in the Half-roll task, the Hand-coded baselines HC-Nav and HC-Forw do not find any solution making the hexapod fall on its back; whereas AURORA discovers such behaviours (see in corners of Half-roll containers in Fig.~\ref{fig:iclr_01_archives}).
In the end, AURORA manages to find containers of behaviours exhibiting diversity in all BD spaces.

The MeS variant, whose BD corresponds to the mean of each stream of sensory data, does not provide the same level of efficiency as AURORA.
In the three tasks, AURORA obtains a coverage that is higher or equal to the coverage obtained for MeS (Fig.~\ref{fig:iclr_02_cov_per_min_fit}).
When considering high minimum performance, the gap between the coverages from AURORA and MeS is even more apparent.
This suggests that the automatic definition of an unsupervised BD may improve the performance upon hand-defined BDs, even if those are calculated considering all full states of the robot.
In the end, by considering the entire full-state trajectory of the agent, AURORA can learn a large diversity of skills in an unsupervised manner, which is more general than any of the compared methods.

\section{Conclusion and Future Work}

In this paper, we provided an additional analysis of AURORA, a QD algorithm that automatically defines its BDs.
In that analysis, we studied what kind of behaviours emerge when AURORA encodes the full-state trajectory of an hexapod robot. 
Our experimental evaluation demonstrated that \name{} autonomously finds diverse types of behaviours such as: navigating to different positions, using its legs in different manners, and performing half-rolls.

In order to limit the computational complexity of AURORA, the container is forced to have a limited size (usually of the order of magnitude of 10{,}000 policies).
This property may prevent the algorithm from exploring in detail all the different types of behaviours in open-ended environments.
For example, it would be interesting to simultaneously obtain a complete behavioural sub-container specialised in half-rolls, and another sub-container specialised in navigation.
This problem has been efficiently addressed by \citet{etcheverry2020imgepholmes} with IMGEP-HOLMES, an algorithm that manages to discover several specialised niches of behaviours.
It would be interesting to study how AURORA could integrate the same mechanisms as IMGEP-HOLMES to handle open-endedness.
%
It would also be relevant to try our approach on more open-ended problems, where the robot can interact with various objects and other robots.

\subsubsection*{Acknowledgments}

This work was supported by the Engineering and Physical Sciences Research Council (EPSRC) grant EP/V006673/1 project REcoVER. 
We thank the members of the Adaptive and Intelligent Robotics Lab for their very valuable comments. 


\bibliography{sample-base}
\bibliographystyle{iclr2022_conference}

\end{document}